\documentclass[preprint,nopreprintline,12pt]{elsarticle}
\usepackage{amssymb}
\usepackage{amsmath}
\usepackage{booktabs}
\usepackage{hyperref}
\usepackage{xcolor}
\hypersetup{
  colorlinks   = true, %Colours links instead of ugly boxes
  urlcolor     = red, %Colour for external hyperlinks
  linkcolor    = blue, %Colour of internal links
  citecolor   = blue %Colour of citations
}
\usepackage{subcaption}
\usepackage{breakurl}
\usepackage{soul}
\usepackage{lineno}
\journal{Nuclear Physics B}

\newcommand{\REV}[1]{#1}%\hl{#1}}
\newcommand{\CB}[1]{#1}%\colorbox{yellow}{#1}}

\begin{document}

\begin{frontmatter}
\title{Automated Thoracolumbar Stump Rib Detection and Analysis in a Large CT Cohort}
 %%%%%%%%%%%%%%%%%%%%%%%%%%%
\author[mri,aimed]{Hendrik Möller}
\affiliation[mri]{organization={Department for Interventional and Diagnostic Neuroradiology, TUM University Hospital},
             addressline={Ismaninger Straße 22},
             city={Munich},
             postcode={81675},
             state={Bavaria},
             country={Germany}}

\affiliation[aimed]{organization={Chair for AI in Healthcare and Medicine},%Department and Organization
            addressline={Technical University of Munich (TUM) and TUM University Hospital}, 
            city={Munich},
            postcode={81675}, 
            state={Bavaria},
            country={Germany}}
\affiliation[rad]{organization={Department of Radiology, Klinikum rechts der Isar, Technical University of Munich},
             addressline={Ismaninger Straße 22},
             city={Munich},
             postcode={81675},
             state={Bavaria},
             country={Germany}}
\affiliation[germancancer]{organization={German Cancer Consortium (DKTK), Munich partner site},
             addressline={},
             city={Heidelberg},
             %postcode={},
             %state={},
             country={Germany}}
\affiliation[zurich]{organization={Department of Quantitative Biomedicine, University of Zurich},
             addressline={Winterthurerstrasse 190},
             city={Zurich},
             postcode={8057},
             state={},
             country={Switzerland}}
\affiliation[imperial]{organization={Department of Computing, Imperial College London},
             addressline={180 Queen's Gate},
             city={London},
             postcode={SW7 2AZ},
             %state={},
             country={England}}
\affiliation[rostock]{organization={Department of Diagnostic and Interventional Radiology, University Medical Center Rostock},
             addressline={Schillingallee 35},
             city={Rostock},
             postcode={18057},
             %state={},
             country={Germany}}
\affiliation[cornell]{organization={Department of Radiology, Weill Cornell Medicine},
             addressline={1300 York Ave},
             city={New York},
             postcode={10065},
             state={New York},
             country={USA}}
\author[rostock,mri]{Hanna Schön} %% Author name
\author[aimed]{Alina Dima} %% Author name
\author[mri]{Benjamin Keinert-Weth} %% Author name
\author[mri,aimed]{Robert Graf} %% Author name
\author[mri,aimed]{Matan Atad} %% Author name
\author[cornell]{Johannes Paetzold} %% Author name
\author[rad,aimed]{Friederike Jungmann} %% Author name
\author[mri,germancancer]{Rickmer Braren} %% Author name
\author[zurich,mri]{Florian Kofler} %% Author name
% Profs
\author[zurich]{Bjoern Menze} %% Author name
\author[aimed,imperial]{Daniel Rueckert} %% Author name
\author[mri]{Jan S. Kirschke} %% Author name
%%%%%%%%%%%%%%%%%%%%%%%%%%%
%% Abstract
\begin{abstract}
%% Text of abstract
%TODO You are required to provide a concise and factual abstract. The abstract should briefly state the purpose of your research, principal results and major conclusions. Some guidelines:
%Abstracts must be able to stand alone as abstracts are often presented separately from the article.
%Avoid references. If any are essential to include, ensure that you cite the author(s) and year(s).
%Avoid non-standard or uncommon abbreviations. If any are essential to include, ensure they are defined within your abstract at first mention.
Thoracolumbar stump ribs are one of the essential indicators of thoracolumbar transitional vertebrae or enumeration anomalies. While some studies manually assess these anomalies and describe the ribs qualitatively, this study aims to automatize thoracolumbar stump rib detection and analyze their morphology quantitatively. To this end, we train a high-resolution deep-learning model for rib segmentation and show significant improvements compared to existing models (Dice score 0.997 vs. 0.779, p-value $< 0.01$). In addition, we use an iterative algorithm and piece-wise linear interpolation to assess the length of the ribs, showing a success rate of 98.2\%. When analyzing morphological features, we show that stump ribs articulate more posteriorly at the vertebrae ($-19.2 \pm 3.8$ vs $-13.8 \pm 2.5$, p-value $<$ 0.01), are thinner ($260.6 \pm 103.4$ vs. $563.6 \pm 127.1$, p-value $<$ 0.01), and are oriented more downwards and sideways within the first centimeters in contrast to full-length ribs. We show that with partially visible ribs, these features can achieve an F1-score of 0.84 in differentiating stump ribs from regular ones. We publish the model weights and masks for public use. % 153 words, 895 chars
% Instead of F1, yield success rate?
\end{abstract}
\begin{keyword}
segmentation \sep ribs \sep thoracolumbar stump rib \sep deep learning \sep computed tomography
%TODO You are required to provide 1 to 7 keywords for indexing purposes. Keywords should be written in English. Please try to avoid keywords consisting of multiple words (using "and" or "of").
%We recommend that you only use abbreviations in keywords if they are firmly established in the field.
%% keywords here, in the form: keyword \sep keyword
%% PACS codes here, in the form: \PACS code \sep code
%% MSC codes here, in the form: \MSC code \sep code
%% or \MSC[2008] code \sep code (2000 is the default)
\end{keyword}
%%%%%%%%%%%%%%%%%%%%%%%%%%%
\end{frontmatter}
%%%%%%%%%%%%%%%%%%%%%%%%%%%
\section{Introduction}
\label{introduction}

%The introduction should clearly state the objectives of your work. We recommend that you provide an adequate background to your work but avoid writing a detailed literature overview or summary of your results.

The human spine typically consists of 24 presacral vertebrae, of which seven are cervical, twelve  rib-bearing thoracic, and five are lumbar vertebrae \cite{thawait2012spine}. Nevertheless, the number of vertebrae may vary and thus lead to thoracolumbar transitional vertebrae (TLTV). TLTV is defined as a vertebra that has partial features of thoracic as well as lumbar vertebrae and has a dysplastic/hypoplastic rib \cite{tatara2021changes, wigh1980thoracolumbar}. These so-called stump ribs, by common definition, have a length of $<= 38$ mm and can only appear at the lowest thoracic level \cite{wigh1980thoracolumbar}. Moreover, Tatara et al. \cite{tatara2021changes} found additional indicators for TLTV, such as anomalies in rib shape. This can occur, for example, due to a deformation of the rib head. 
The thoracolumbar junction is considered a weak connection, so injuries, such as fractures or thoracolumbar junction syndrome, often occur here \cite{kim2015treatment, du2018differentiation}. These changes in biomechanics might be crucial for surgical planning, as they can affect surgical outcome and recovery \cite{poolman2023thoracolumbar, mahato2018thoracolumbar}. Additionally, the occurrence of TLTV often indicates numerical anomalies, such as an additional or missing thoracic vertebra. 

Overall, detecting such anomalies is vital for the correct assessment of vertebrae heights and localizing weak points. Automating this process would reduce radiologists' work and increase the accuracy of surgical planning. Moreover, some CT reformations are not wide enough to capture the whole extent of the ribs. In these cases, a manual assessment of thoracolumbar stump ribs is so far infeasible. We aim to automatically localize and detect stump ribs and exploit rip-shape differences between stump ribs and regular ribs to detect stump ribs even with partially visible ribs. 

For this to work, we need to be able to automatically localize ribs in a CT image, measure their length, and extract morphological features. We opt to solve this by creating dense and complete segmentation masks. \REV{Segmentation is the task of assigning a label to each pixel of an input image} \cite{csurka2022semantic}\REV{. In our case, the input images are CT images and we want to assign to each voxel whether it belongs to a rib structure.} We then calculate the length of the ribs in addition to morphological features from these segmentations. After that, we can compare stump ribs and regular ribs in terms of those features, and use these features to differentiate the two groups without the use of the length.

For developing solutions to automate the creation of segmentation annotations, deep learning has become a prominent field in science. Specifically, the task of segmentation has been used in the past for localizing and describing medical structures \cite{isensee2021nnunet} and is considered state-of-the-art. Previously, most existing approaches related to ribs utilized segmentation or center extraction algorithms \cite{shen2004tracing, klinder2007segrib, wu2021segrib, staal2007segrib, lenga2019segribcl, wu2012segribtemplatecl, wang2020mdu, ramakrishnan2011ribcl}, which are not publicly available. The TotalSegmentator \cite{wasserthal2023totalsegmentator} is publicly available and can segment ribs, yet it is unable to yield a quality sufficient for rib-length calculations. We expand the rib segmentation to capture the entire rib, develop a length measurement algorithm, and make our model and segmentation masks available, thus bridging a gap in the community. Finally, to the best of our knowledge, there has been no work trying to automatize stump rib detection or quantitative experiments contrasting regular ribs with stump ribs using automated morphological features. 

The contributions of this study are:
\begin{enumerate}
    \item We develop a whole rib segmentation tool for CT imaging and present a rib length measurement algorithm based on that segmentation.
    \item We automatically detect and localize thoracolumbar stump ribs by calculating the rib length.
    \item We describe stump rib morphological features in contrast to regular ribs and propose them as additional indicators of stump rib occurrence.
    \item We show that these morphological features classify stump ribs even with a partial field of view.
    \item We make the segmentation model, assessment algorithm, and manual and predicted rib segmentation masks publicly available (\burl{https://github.com/Hendrik-code/rib-segmentation}).
\end{enumerate}
\section{Materials and methods}
\label{methods}

%The materials and methods section should provide sufficient details about your materials and methods to allow your work to be reproduced by an independent researcher. Some guidelines:
%If the method you used has already been published, provide a summary and reference the originally published method.
%If you are quoting directly from a previously published method, use quotation marks and cite the source.
%Describe any modifications that you have made to existing methods.

\subsection{Data}

\begin{figure}
    \centering
    \includegraphics[width=\textwidth]{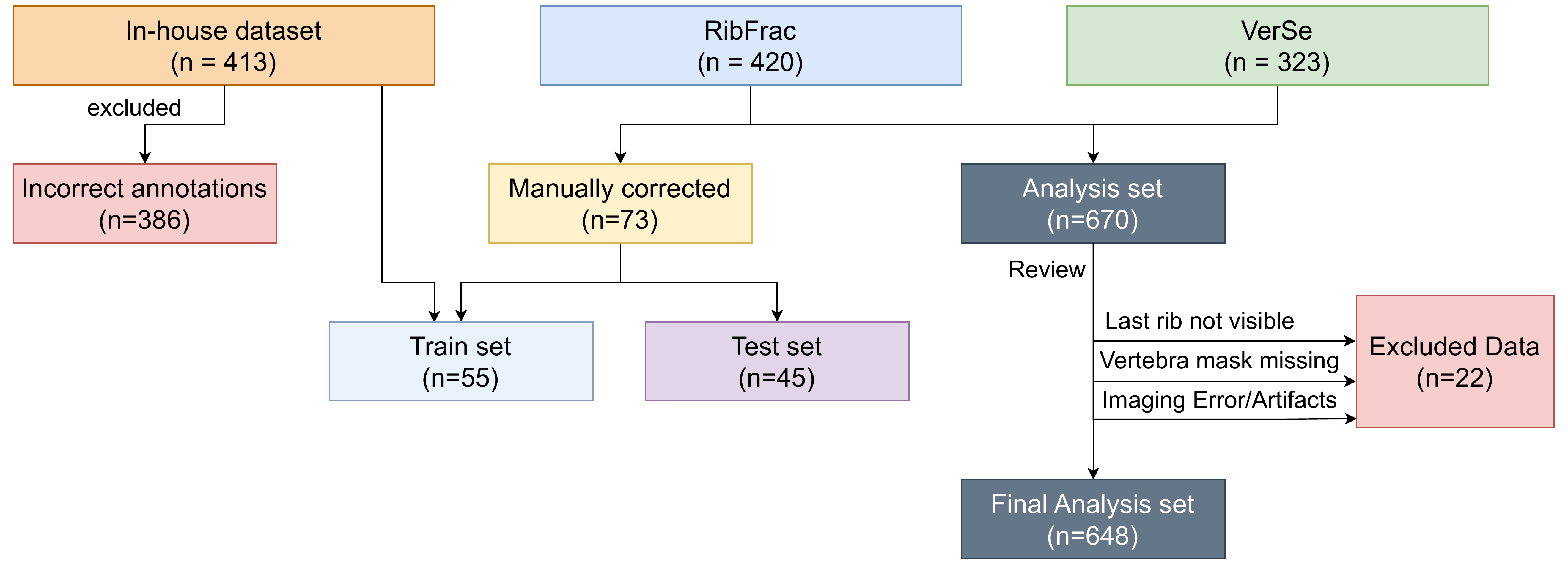}
    \caption{The flow of our three utilized datasets. We used the private in-house dataset for training the rib segmentation \REV{only} and performed all downstream experiments on public data only.}
    \label{fig:dataflow}
\end{figure}

This study used three CT datasets (see Figure \ref{fig:dataflow}). Two publicly available ones, VerSe \cite{sekuboyina2021verse} and RibFrac \cite{ribfrac2020, yang2024ribfrac}, and a private in-house dataset (waived informed consent by the local ethics committee, 27/19 S-SR dated 22nd April of 2020). For demographics, see Table \ref{tab:demo}. The VerSe data has expert-quality vertebra segmentation annotations. Thanks to Jin et al. \cite{jin2023ribseg}, the RibFrac data comes with instance segmentations for the ribs (called RibSeg). The in-house dataset has binary annotations for the ribs, separating rib tissue and background. We \REV{selected} 27 subjects from the in-house dataset \REV{with very-high segmentation quality} (see Figure \ref{fig:dataflow}).

\subsection{Segmentation Approach}
\REV{To measure the length of the ribs, we require dense and complete segmentation annotations. We define a correct rib annotation as the filled bone outline (indicated by high-intensity values in CT) up to, but not including, the costal cartilage and the sternum.}

\begin{figure}[!htbp]
    \includegraphics[width=\textwidth]{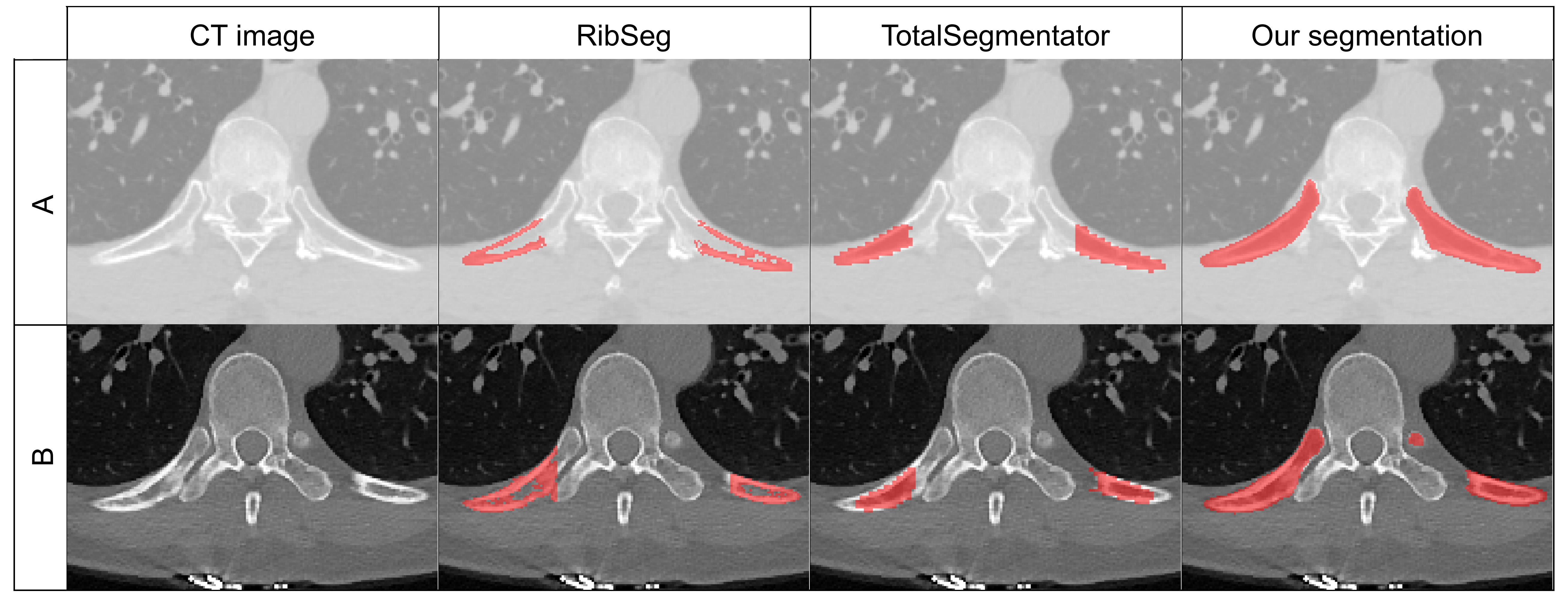}
    \caption{Two example subjects (rows) of the RibFrac dataset. We compare the original RibSeg annotation with the prediction of the TotalSegmentator and our rib segmentation model. Both the RibSeg annotation and the one from TotalSegmentator fail to annotate the rib part close to the vertebra. Furthermore, the RibSeg annotation is not dense but contains only the outline, and the one from TotalSegmentator lacks detail as it segments with an isotropic resolution of 1.5 mm. Our model is trained for 0.8 mm isotropic space; thus, it produces smooth and high-resolution annotations. This behavior is true for all samples in our datasets.}
    \label{fig:ribseg}
\end{figure}

We observed that the rib annotations from the RibFrac dataset exclude the head of the ribs. The TotalSegmentator \cite{wasserthal2023totalsegmentator} exhibits the same issue (see Figure \ref{fig:ribseg}). Thus, we developed our own rib segmentation model by utilizing nnUNet \cite{isensee2021nnunet}, a widely used architecture for training semantic segmentation models. We started with a small set of annotations from the in-house data, ran inference on the public datasets, manually corrected some to expand the training set, and then trained one more time.

In detail, we used 27 subjects from the in-house data with high segmentation quality. We trained a default nnUNet with suggested parameters with the exception of the patch size, which we set to a consistent cubic size of 192 x 192 x 192. We trained for 300 epochs with 3-fold cross-validation, resampling all training data to 0.8 mm isotropic space. Additionally, we utilized elastic deformation and random horizontal flipping to augment our data tenfold. 

We then applied this model to the public datasets. After careful review, we chose 17 subjects from VerSe and 11 from RibFrac.
These annotations and another 45 randomly chosen subjects from VerSe and RibFrac were manually corrected by an expert (H.M.) with two years of experience supervised by two experts (H.S., J.S.K.) of three and 23 years of experience, respectively. We used ITKSnap 3.8.0 \cite{yushkevich2006itksnap} for annotation review and manual correction. The 45 randomly chosen subjects were exclusively used as test set.
The resulting training split totaling 55 subjects was used to retrain the rib segmentation model using the same training procedure. 

Figure \ref{fig:ribseg} compares our rib segmentation tool to existing approaches. These rib annotations are binary and thus only distinguish between foreground (ribs) and background (not ribs). The created manual and predicted segmentation masks from this study are published under \burl{https://doi.org/10.5281/zenodo.14850928}.

\subsection{Instance Assignment}

\begin{figure}
    \centering
    \includegraphics[width=\textwidth]{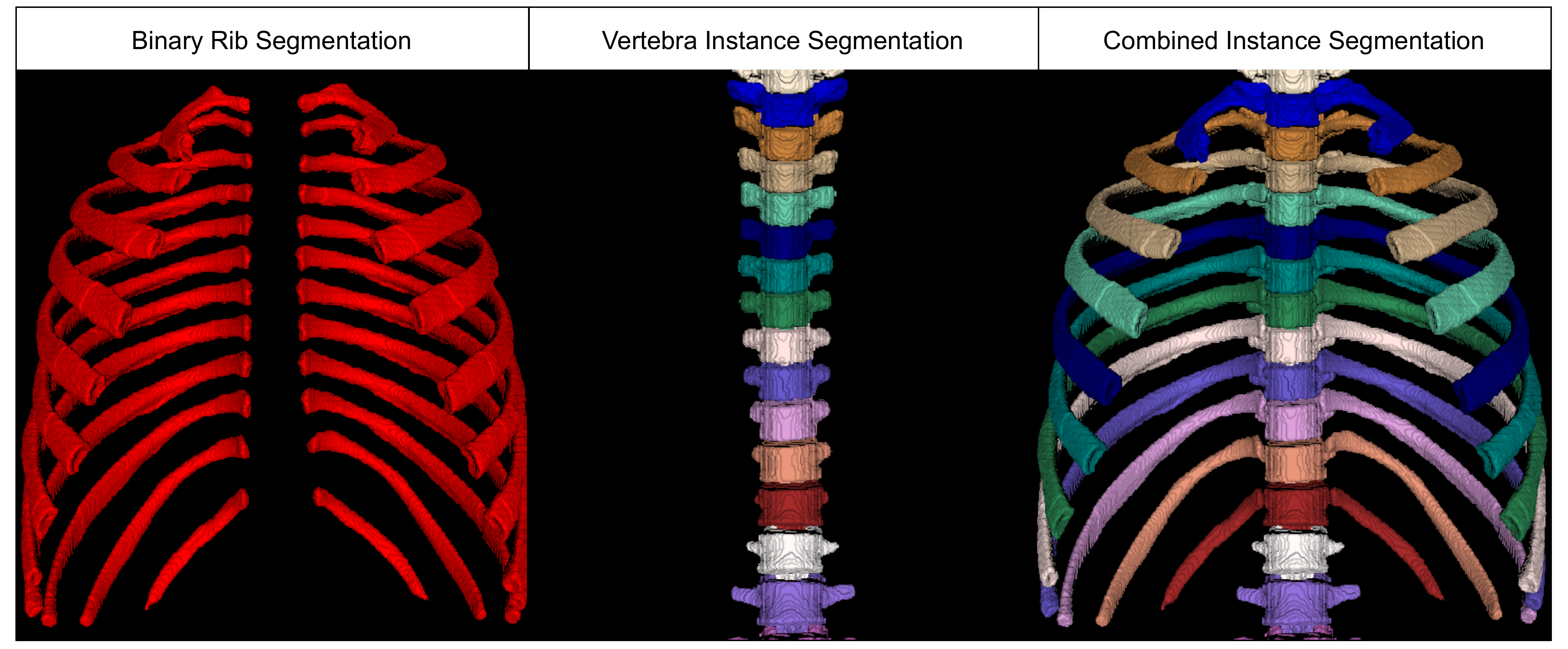}
    \caption{An example showing our transition from semantic rib segmentation to instance rib segmentation by using the vertebra instance segmentation provided by SpineR. From left to right: Binary rib segmentation; vertebra instance segmentation; resulting combined instance annotation.}
    \label{fig:ribinstance}
\end{figure}

Our goal is to derive information for each rib individually. This requires instance segmentation masks of the ribs, not only semantic ones. To bridge this gap, we used vertebra instance segmentation masks. The VerSe data already contained expert-level annotations for individual vertebrae. We used the Bonescreen SpineR tool (Bonescreen GmbH, Bavaria, Munich) based on Sekuboyina et al. \cite{sekuboyina2018btrfly} to generate vertebra segmentations for the RibFrac dataset. We then derived the connected components from the predicted binary rib segmentation mask of our model and assigned the rib components to the vertebrae based on spatial proximity. To ensure robust results, we employed the constraints that each vertebra can only have two ribs, and each rib can only be assigned to one vertebra (see Figure \ref{fig:ribinstance}). Moreover, we distinguished the side of the rib by comparing the rib localization to its corresponding vertebra. We encountered instances where a rib was segmented, but its corresponding vertebrae had not been segmented. In these cases, we consistently relabeled these rib instances to a set of unused labels. We manually reviewed the subjects using generated 2D coronal and sagittal snapshots.

\subsection{Rib Length Measurement}

Now that we can localize and view each rib in our images individually, we need to measure the length of a single rib to assess whether it is a stump rib (SR). For that, we employed an iterative linear algorithm calculating points on a path.
We started by cropping a single rib combined with its corresponding vertebra. To ensure consistency, we rescaled the cropped segmentation mask to a fixed resolution of 0.5 mm for each dimension. We filled small holes to ensure the segmentation was dense while keeping the overall shape and size consistent.

\begin{figure}[!htbp]
    \centering
    \includegraphics[width=0.99\textwidth]{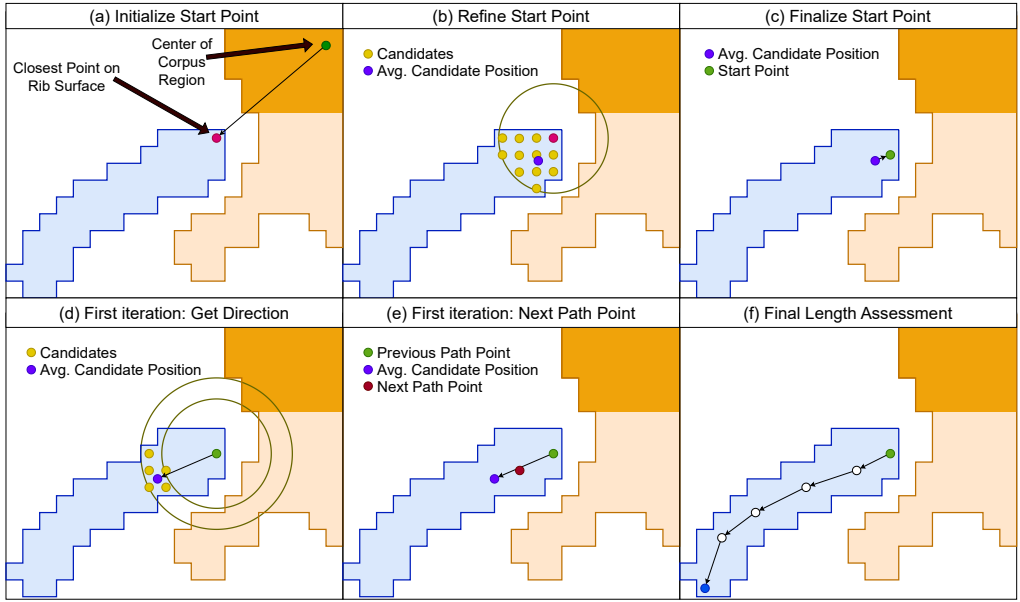}
    \caption{2D Schematic illustration of our 3D rib length measurement algorithm (RLMA). The blue contour represents the rib, while the corresponding vertebra is orange (the vertebral corpus is darker). The first row shows the initialization of the starting point. First, we find the closest point of the rib surface to the center of the corresponding vertebra corpus (a); second, we use the surrounding points to find an average starting position (b), and third, we finalize the starting point by projecting back to the rib surface (c). The second row shows an iteration: Finding candidate points by looking at a fixed circular distance (d), then taking the average of those candidate points and finding the next point on the path by using the direction vector of the previous point and the average of candidate points (e). The last panel (f) shows the example after the RLMA has calculated all path points. The figure demonstrates the algorithm in two dimensions for clarification purposes, while the RLMA works in three-dimensional data.}
    \label{fig:riblength}
\end{figure}

The rib-length measuring algorithm (RLMA) starts by finding the start point of the rib. We select the closest point on the segmentation surface to the center of mass of the corresponding vertebra corpus. We refine that location by taking the average location of all surrounding segmented voxels and projecting this onto the segmentation surface (see Figure \ref{fig:riblength}). This ensures that the start position is already at the middle of the rib, not at a corner position. We set this start point as the first path point. 

From the start point, we iteratively add path points along the center of the rib. For this, we calculate all candidate points within a circular distance of 14.5 to 15.5 mm from the previous path point. We remove all candidate points in that list that are closer to any prior path point than the latest, ensuring that we are iterating in the correct direction. We then take the average coordinate location of all these candidate points. We can compute a direction vector in relation to our previous path point. Moving from the latest point halfway along that direction vector gives us our next target. This new location gets projected onto the nearest segmentation voxel and is added as the next path point (see Figure \ref{fig:riblength}). This gets repeated until we reach the end of the rib, where no possible points lie on the circular slice. To find the final path point, we shoot a cone of ray-casts from the latest path point along the vector of the previous path point and find the point in those ray-casts that are furthest away from the latest path point. We take this as the end path point.
In order to calculate the length, we take the points on the path and sum the distances between each path point.

By definition, thoracolumbar stump ribs (SR) are shorter than or equal to 38 mm \cite{wigh1980thoracolumbar}. We compare this threshold to the length computed by our RLMA. This yields a binary label (SR or not-SR) for each rib.

\subsection{Morphological Features}

\begin{figure}[!htbp]
    \centering
    \includegraphics[width=0.6\textwidth]{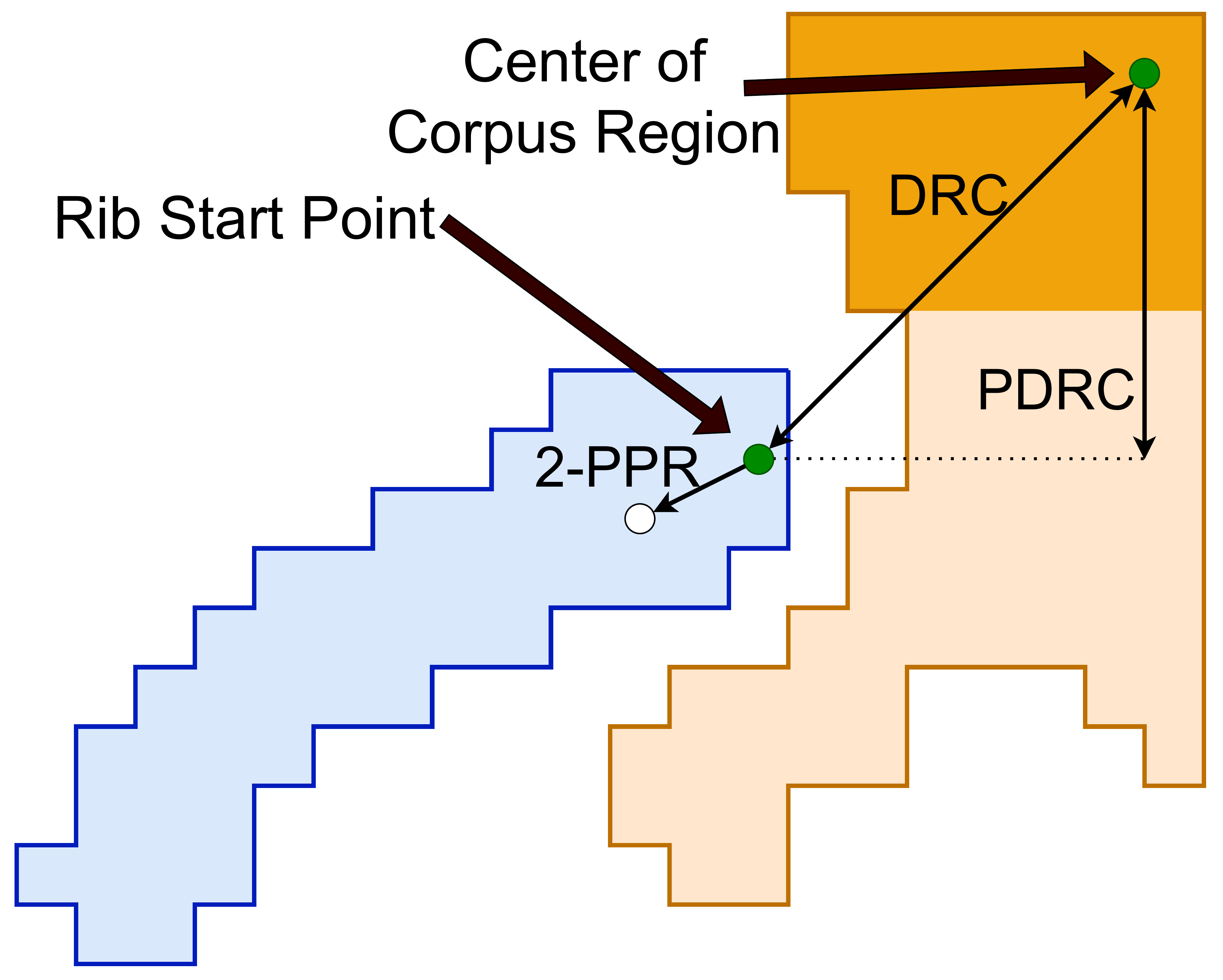}
    \caption{Example 2D image to showcase some of our calculated 3D features. Blue is the rib, and orange is its corresponding vertebra. DRC is the spatial distance and direction between the rib start point and the center of the vertebra corpus region (dark orange). PDRC is the posterior component of the DRC. $2$-PPR: The spatial relation for the first two path points (i.e., the direction vector between the start point and the next path point).}
    \label{fig:ribfeatures}
\end{figure}

We calculate morphological features from our rib segmentations. We compute the spatial relation between the rib's start point and the corresponding vertebra corpus center (DRC). Specifically, we look at the posterior distance between the rib and corpus center, which we denote as PDRC (see Figure \ref{fig:ribfeatures}). Additionally, we use the path points and their spatial relation to each other using the direction vectors between neighboring path points (PPR). For the PPR, we denote $n$-PPR as using the first $n$ path points relations. From the RLMA, we have the rib's length; using the segmentation mask, we can also calculate the volume-to-length ratio.

To account for rotation, we take the orientation of the respective vertebra and multiply the inverted rotation matrix on the measured features to get rotations relative to the vertebra. Additionally, to be consistent, we mirror the left ribs as if they were on the right side. This means that the right dimension always points away from the vertebra.

\subsection{Experiments}

We utilized the two public datasets (VerSe and RibFrac) for all downstream experiments. We ran our rib segmentation across all subjects and the rib measurement algorithm (RLMA) over the lowest two vertebrae with ribs present.

We compare the segmentation performance on our manually corrected test set. Our baselines are the RibSeg segmentation masks (both with and without fill holes applied) and the TotalSegmentator. \REV{We evaluate this on various metrics (see }\ref{sec:metrics}\REV{).}
This yields a comparative analysis regarding the segmentation performance. 

For the rib length analysis, we used the analysis set (all subjects from VerSe and RibFrac that are not in the training set) and incorporated some exclusion criteria: We excluded 36 subjects in which the last rib is not visible in the image, 78 ribs for which the corresponding vertebra segmentation was missing, and two subjects with imaging artifacts (see Figure \ref{fig:dataflow} for detailed data flow).

Then, two experts (H.S., B.K.) reviewed the generated ribs segmentation for segmentation and measurement errors. As there is no standard in manually measuring the rib length, the experts categorized each rib into a five-point Likert scale: (1) the segmentation has major errors or segmented the wrong structure, (2) the segmentation is the correct structure but erroneous enough so that the resulting length is wrong, (3) the segmentation touches the border of the image and thus the length of the visible section is correct but incomplete, (4) the segmentation has minor mistakes which did not impact the length calculation, and (5) everything is fine.
The relative frequency of this rating indicates how often our segmentation, combined with our RLMA, succeeded in measuring the rib length.

For the morphological analysis part, we included the ribs with certainty about the SR label. Hence, we only used ribs marked with (4) and (5) and those with (3) where the length of the visible part already exceeded the SR threshold.

We show how well we can separate stump ribs from regular ribs using morphological relations to the corresponding vertebra without using the full length of the ribs. Thus, we use the DRC features as input for a Support Vector Machine (SVM) \cite{cortes1995support} to differentiate stump ribs and regular ribs. We also used $n$-PPR as input features for the SVM. We test this with $n$ as 2, 3, and 4 to see how much of a rib needs to be visible in order to predict stump ribs. Going beyond $n=4$ would result in a measured length that is close to or beyond the SR threshold, thus defeating the purpose. We also combine $n$-PPR and DRC as input features to see whether this improves the performance.

For all SVM experiments, we repeat the setup with 10 different fixed seeds and a random 70/30 train/test subject-wise split to assess a robust separability performance. We report the F1-score averaged over the 10 runs. Additionally, we experimented with a polynomial kernel compared to a linear one.

\subsection{\CB{Evaluation Metrics}}\label{sec:metrics}

\REV{We evaluated the rib segmentations produced by our model using multiple metrics.}

\REV{Dice similarity coefficient (DSC): It is a crucial overlap score that measures how many voxels between the prediction and the reference annotation match. With $X$ and $Y$ being the predicted and reference annotations, respectively, we can calculate the DSC with Equation 1.}

\begin{align}
    \text{DSC}(X,Y) = \frac{2 |X \cap Y|}{|X| + |Y|} \in [0,1]
\end{align}

\REV{Average symmetric surface distance (ASSD): This metric reports the average of all the distances from points on the boundary of the predicted annotation to the boundary of the reference annotation, and vice versa. With $\text{d}(a,b)$ being the distance between two points $a$ and $b$, and $A$, $B$ being the boundaries of $X$ and $Y$, respectively, we can calculate the ASSD via Equation 2. ASSD is important to show how far away erroneous or missing voxels are from the reference annotation, i.e., whether errors are local.}
\begin{align}
     \text{ASSD}(X,Y) &= \frac{\text{asd}(A,B) + \text{asd}(B,A)}{ |X| + |Y|} \in [0,\infty] \\
     \text{asd}(A,B) &= \sum_{a \in A}\min_{b \in B}\text{d}(a, b) \in [0,\infty]
\end{align}

\REV{Utilizing panoptica} \cite{kofler2023panoptica} \REV{, we match instances with a DSC greater than or equal to $0.5$ as true positives. With that in mind, we calculate the following instance-wise metrics:}

\REV{Recognition Quality (RQ): The F1-score of instance detection based on the above definition of true positives. With tp, fp, and fn being true positives, false positives, and false negatives, respectively, we calculate the RQ via Equation 4. This score indicates the detection quality of the ribs.}

\begin{align}
    \text{RQ}(\text{tp},\text{fp}, \text{fn}) = \frac{\text{tp}}{\text{tp} + \frac{1}{2} \cdot (\text{fp} + \text{fn})} \in [0,1]
\end{align}

\REV{With a metric $M$, we define the segmentation quality ($\text{SQ}_{M}$) as the average of the metric $M$ across all true positives, defined by Equation 5. This yields how well the detected instances have been segmented.}

\begin{align}
    \text{SQ}_M(\text{tp}) = \sum_{(i_{\text{ref}}, i_{\text{pred}}) \in \text{tp}} \frac{M(i_{\text{ref}}, i_{\text{pred}})}{|\text{tp}|}
\end{align}

\REV{Finally, the panoptic quality (PQ) combines the RQ and SQ. Again using a metric $M$, we define the PQ with Equation 6.}

\begin{align}
    \text{PQ}_M(X,Y) = \text{SQ}_M \cdot \text{RQ}
\end{align}

\REV{We evaluate the rib segmentation model using the DSC on the whole masks, i.e., calculating a binary overlap score (binary DSC), as well as the instance-wise metrics $\text{RQ}$, $\text{SQ}_{\text{DSC}}$, $\text{PQ}_{\text{DSC}}$, and $\text{SQ}_{\text{ASSD}}$ for a detailed comparison.}

%\REV{We evaluated the rib segmentation model using the Dice similarity coefficient (DSC) and the average symmetric surface distance (ASSD). We utilize panoptica} \cite{kofler2023panoptica} \REV{to derive instance-wise metrics -- Recognition Quality (RQ), Segmentation Quality (SQ) and Panoptic Quality (PQ). Instances with an intersection over union (IoU) greater than or equal to $0.5$ were considered true positives.}

\subsection{Statisticial Analysis}

To compare the statistical significance of our segmentation models, we use the Wilcoxon signed rank test \cite{Rey2011} on Dice metrics, with $p < 0.05$ indicating statistical significance. For the feature distributions grouped in stump ribs and regular ribs, we calculate the Wilcoxon rank sum test and use the same threshold for statistical significance.
\section{Results and Discussion}
\label{results}

%Results should be clear and concise. We advise you to read the sections in this guide on supplying tables, artwork, supplementary material and sharing research data.

\subsection{Segmentation Performance}

Our model trained on 55 samples outperforms the TotalSegmentator on 45 randomly selected, manually corrected test subjects (see Table \ref{tab:evaltest}). We hypothesize this is primarily due to the TotalSegmentor not being able to segment the beginning part of the ribs and its inferior segmentation resolution. 

The public RibSeg annotations \cite{jin2023ribseg} for the RibFrac dataset are also inferior compared to our rib segmentation model (see Table \ref{tab:evaltestribfrac}). Our model consistently outperforms on every metric and sample with a significant margin (p-value $< 0.01$).
This performance indicates that we can automatically and consistently segment the whole ribs in arbitrary field-of-views. %We publish the model weights and the manually annotated rib segmentations, thus enabling others to create whole rib segmentations.

\subsection{Rib Length Measurement}

Based on the manual review of our experts, 98.2 \% of the lowest two ribs in both public data sets were correctly segmented and measured (see Table \ref{tab:testdata}), thus yielding a correct stump rib assessment. Only in 1.8\% of the cases did we have segmentation errors that influenced the rib length measurement.
After a manual review of these cases, there was only one rib with a measured length close to the stump rib threshold of 38 mm. We observed that the length calculation is only erroneous in a minor way \REV{for all other cases}. Thus, we can confidently \REV{state} that the errors in rib length calculation for all but one of these cases would not have influenced their stump rib label. \REV{For a qualitative analysis of the observed errors, see }\ref{app:failure}\REV{.}

This shows that our approach can robustly segment ribs and measure their length in CT images, thus automatically detecting and localizing stum\REV{p} ribs based on the generally accepted, length-based definition.

\subsection{Morphological Features}

After our exclusions, we ended up with 2464 individual ribs across 648 subjects, all either the lowest or second-lowest rib of the subject. We did not observe any bias in our exclusion criteria, such as pathologies or anomalous intensities. Of the 2464 ribs that we analyzed, 159 were stump ribs. Of the 648 subjects, 133 had at least one SR (20.5\%), according to the 38mm length threshold for stump ribs. 

\begin{figure}[!htbp]
    \centering
    \includegraphics[width=0.65\textwidth]{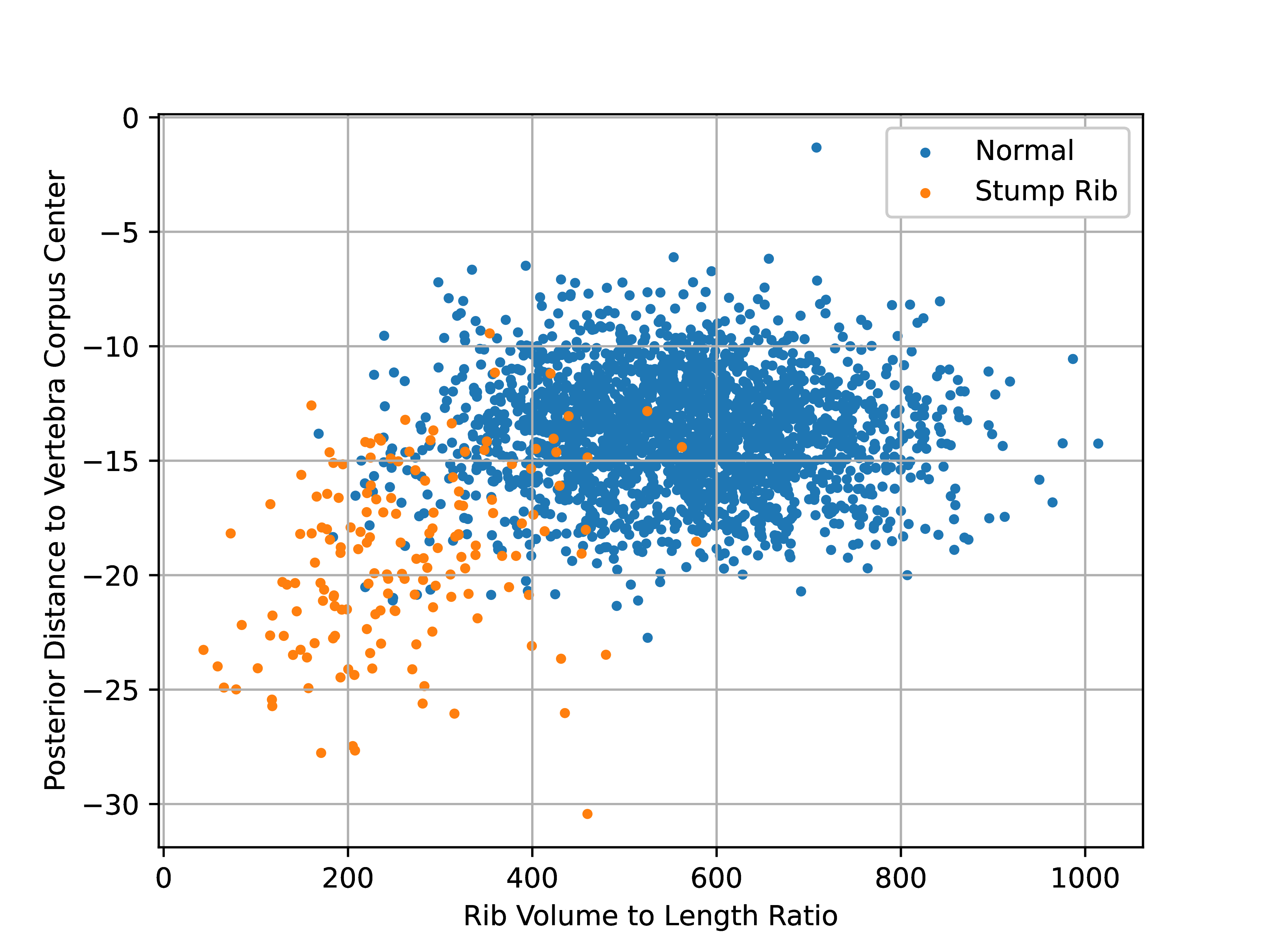}
    \caption{Analyzed ribs classes are colored as follows: SR (orange) and regular rib (blue). The x-axis shows the volume of a rib divided by its length, and the y-axis shows the PDRC. Although not separated, we see a clear trend that stump ribs have a lower volume-to-length ratio and a greater posterior distance between the corpus and the rib, i.e., the rib starts further back.}
    \label{fig:49dist_volumebylength}
\end{figure}

On average, the distance in the posterior direction between the corresponding vertebra corpus center and the closest point on the rib was increased for stump ribs ($-19.2 \pm 3.8$ vs $-13.8 \pm 2.5$, p-value $<$ 0.01), and they had a lower volume-to-length ratio ($260.6 \pm 103.4$ vs. $563.6 \pm 127.1$, p-value $<$ 0.01, see Figure \ref{fig:49dist_volumebylength}).

We manually inspected the samples with a high posterior distance to the corresponding corpus, but they were not classified as stump ribs based on the length. Here, we observed that most of the samples' vertebrae had an enlarged fovea pushing the beginning of the rib in the posterior direction or an anomaly in the corpus region, changing the center of the corpus region to an unnatural position. This indicates that the higher posterior distance is not due to the morphology of the rib but rather to that of the vertebra. Thus, we hypothesize that taking the vertebra morphology into account should further enhance the separability between SR and regular ribs, using the posterior vertebra distance relation. 

\begin{figure}[!htbp]%
    \centering
    \begin{subfigure}{0.5\textwidth}%
    \includegraphics[width=0.98\linewidth]{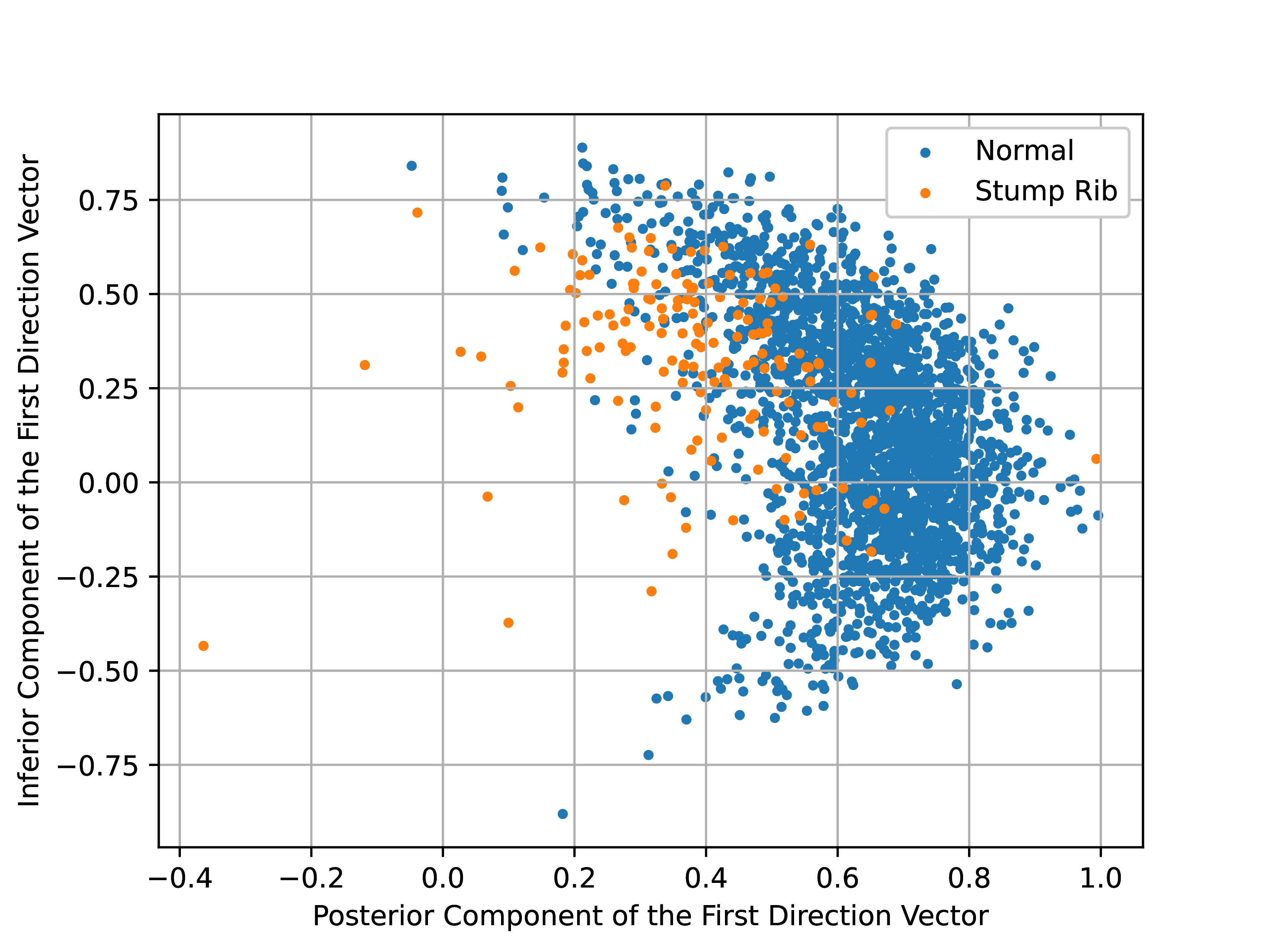}
    %\caption{TBD.}
    \label{fig:pathdirection1}
    \end{subfigure}%
    \begin{subfigure}{0.5\textwidth}%
    \includegraphics[width=0.98\linewidth]{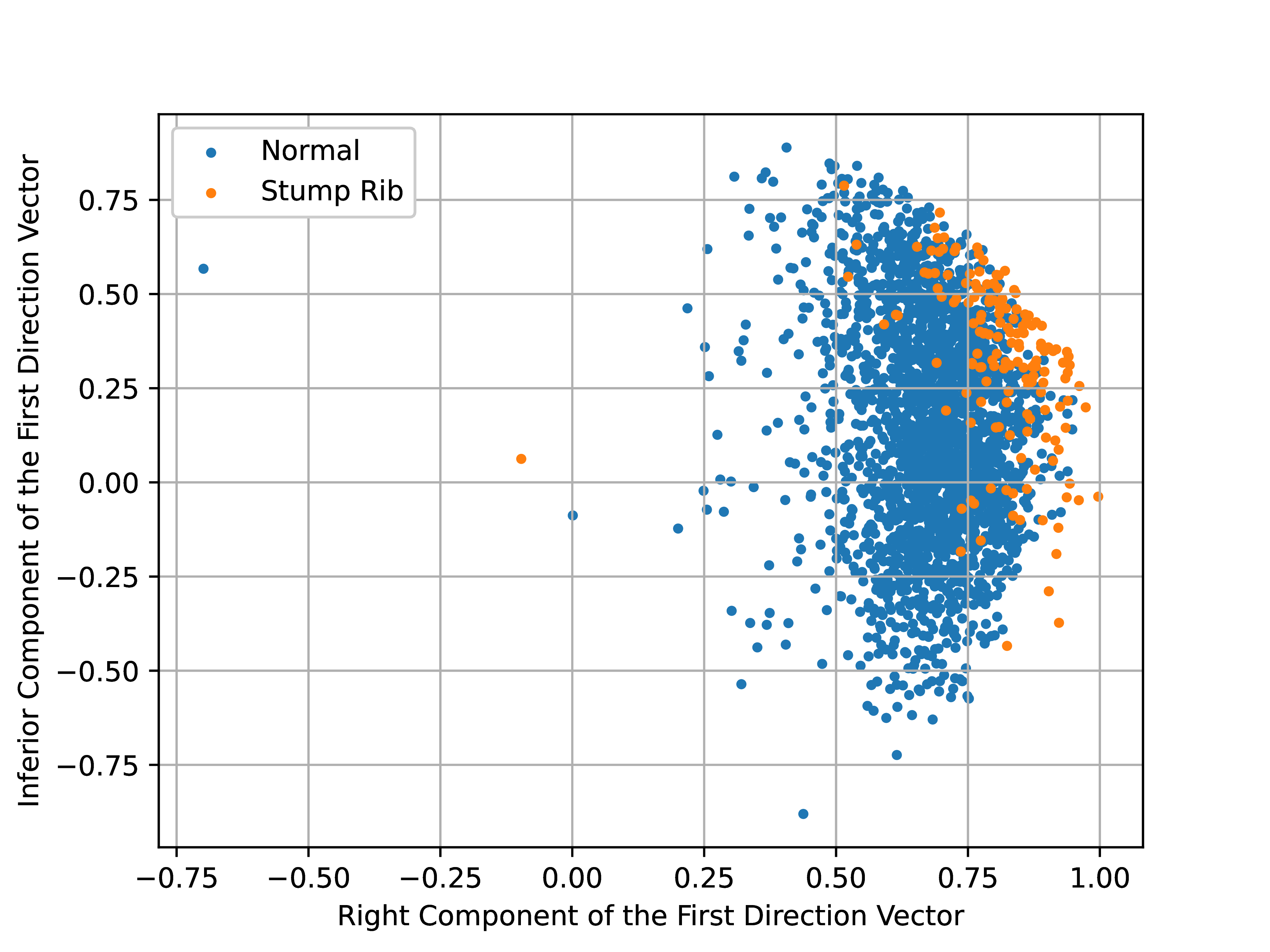}
    %\caption{TBD.}
    \label{fig:pathdirection2}
    \end{subfigure}%
    \caption{Our analyzed rib classes are colored as follows: SR (orange) and regular rib (blue). The two axes show the different components of the vector between the starting point of the rib and its first path point, normalized as a unit vector. We contrast the posterior to the inferior component (left) and the right and inferior component (right).
    }
    \label{fig:pathdirection}
\end{figure}%
When observing the initial orientation of a rib by using the unit vector from the start point to the second path point (equivalent to $2$-PPR, which is roughly the beginning 7 mm of a rib), we see significant differences. The posterior component of stump ribs is smaller ($0.39 \pm 0.17$ vs $0.64 \pm 0.13$, p-value $<$ 0.01), while the inferior component is higher ($0.32 \pm 0.23$ vs. $0.15 \pm 0.30$, p-value $<$ 0.01). Additionally, stump ribs are more lateral angled (right component of $2$-PPR $0.81 \pm 0.11$ vs $0.68 \pm 0.11$, p-value $<$ 0.01). 
In summary, even in the first ~7 mm of a rib, full-length ribs have a more prominent posterior component, meaning the ribs are angled more towards the back of the vertebra at the beginning. In contrast, stump ribs are more outwards and downwards oriented (see Figure \ref{fig:pathdirection}). 

Using the $4$-PPR on each rib, a Support Vector Machine (SVM), trained with a polynomial kernel of degree 5 and random 70/30 splits, reaches an average F1-score of $0.82 \pm 0.04$ across 10 runs in distinguishing SR from non-SR (see Table \ref{tab:svmresults}).
Taking only the first three points ($3$-PPR) reduces the F1-score to $0.7 \pm 0.04$.
Using the relation of the rib start to the vertebra corpus surface and center results in an F1-score performance of $0.62 \pm 0.05$ using the same SVM setup. 
Although DRC alone as input does not perform as well, combining it with PPR mainly boosts the performance (see Table \ref{tab:svmresults}).

We achieved the best performance of $0.84$ by using both $4$-PPR and DRC as inputs to the SVM and using a linear kernel. Notably, to achieve this, we only require the beginning part of the rib and the corresponding vertebra segmented.
This is fascinating, as the first four path points have an average total length of around 22 mm. By looking at the initial $~22$ mm of a rib's direction and orientation, we can deduce whether the rib is a stump rib or not. If only the initial $~15$ mm (required for $3$-PPR) of a rib is visible in the scan, we lose some performance but still get decent results. This is especially interesting in cases where the scan is not very wide, and thus, ribs are not fully visible.
Thus, our experiments imply that ribs could be categorized by morphological features, adding more depth to the classification than a simple cutoff based on the length of the rib.

For a study that analyzes the relation of the stump rib length threshold to the SVM performance, see \ref{app:srthresh}.

\subsection{Limitations}

As there is no gold standard for manually calculating the rib length, we cannot perform a comparative quantitative analysis regarding the exact measurements. Furthermore, the two public datasets VerSe and RibFrac used in this study have backgrounds in spinal anomalies such as fractures and rib cage fractures, respectively. Thus, they do not represent the general population, and we forfeit meaningful normative values or more generalized findings about ribs. Compared to the literature, we have a high number of stump ribs in our data.

\REV{The accuracy of the measured rib length depends on the quality of the produced segmentation masks. If the segmentation is erroneous, so is the measured rib length. However, our results and observations show that only major segmentation errors lead to notable measured rib length errors.}

Finally, although we manually corrected all ribs and evaluated the segmentation performance on all the ribs in a scan, our downstream experiments were focused on the lowest two ribs. Thus, we did not manually review the measurements of upper thoracic ribs and cannot guarantee the same length measuring performance there. However, as only the lowest ribs in a subject can be stump ribs, the measurement of the length is more important there.

We leave it up to future research to include the morphology of the vertebra in a rib-focused analysis, taking abnormalities such as enlarged foveas into account. Additionally, a more extensive population study using our segmentation and rib length measurement tools could further improve our understanding of stump ribs. Moreover, the occurrence of our stump rib features and prevalence related to other aberrations, such as lumbosacral anomalies, should be investigated.
\REV{Lastly, we leave it as an open question whether the rib length measurement algorithm and the morphological features can be inversely utilized to detect rib segmentation errors.}

%%%%%%%%%%%%%%%%%%%%%%%%%%%%%%
%The discussion section should explore the significance of your results but not repeat them. You may combine your results and discussion sections into one section, if appropriate. We recommend that you avoid the use of extensive citations and discussion of published literature in the discussion section.

%The conclusion section should present the main conclusions of your study. You may have a stand-alone conclusions section or include your conclusions in a subsection of your discussion or results and discussion section.

\subsection{Conclusion}

Overall, our proposed method can robustly detect, localize, segment, and measure ribs in CT scans. This allows an automatic classification of stump ribs. Additionally, we propose simple morphological features beyond the length to correctly classify thoracolumbar stump ribs. These enable the detection of stump ribs in limited, arbitrary field-of-view CT images and might help to label vertebrae correctly and improve surgical planning.

Finally, we make the code and model weights publicly available (\burl{https://github.com/Hendrik-code/rib-segmentation}), thus making it easy for fellow researchers to create rib segmentation masks for their own data. Further, we also release the manually corrected and predicted rib masks of the two public datasets: VerSe and RibFrac (\burl{https://doi.org/10.5281/zenodo.14850928}).
\section{Tables with captions}
\label{tables}

\begin{table}[hbpt]
    \centering
    \small
    \begin{tabular}{lccc}
        \toprule
        & VerSe \cite{sekuboyina2021verse} & RibFrac \cite{ribfrac2020} & In-house data \\
        \midrule
        Subjects & 323 & 420 & 143\\
        Sex (\% female) & N/A & 36 & 50\\
        Age range (yrs) & 18 -- N/A & 21 -- 94 & N/A\\
        Mean age (yrs) $\pm$ SD & $59 \pm 17$ & $55 \pm 12$ & $69 \pm 9$\\
        Date range (yrs) & 2013 -- 2020 & N/A & 2016 -- 2020 \\
        Background & spinal anomalies & fractured ribs & pre-treatment CT \\
        \bottomrule
    \end{tabular}
    \caption{The data demographics of this study. We used the public VerSe and RibFrac data and a private in-house dataset.}
    \label{tab:demo}
\end{table}

\begin{table}[hbpt]
    \centering
    \small
    \begin{tabular}{rcc}
        \toprule
        & TS \cite{wasserthal2023totalsegmentator} & Ours \\
        \midrule
        Binary DSC $\uparrow$                & $0.751 \pm 0.078$ & $\textbf{0.997} \pm 0.006$ \\
        $\text{RQ} \uparrow$                 & $0.948 \pm 0.054$ & $\textbf{0.987} \pm 0.053$\\
        $\text{SQ}_{\text{DSC}} \uparrow$    & $0.738 \pm 0.076$ & $\textbf{0.984} \pm 0.043$   \\
        $\text{PQ}_{\text{DSC}} \uparrow$    & $0.700 \pm 0.080$ & $\textbf{0.971} \pm 0.050$\\
        $\text{SQ}_{\text{ASSD}} \downarrow$ & $1.896 \pm 1.167$ & $\textbf{0.064} \pm 0.126$\\
        \bottomrule
    \end{tabular}
    \caption{Evaluation of 45 manually corrected test subjects from the RibFrac and VerSe dataset on the rib segmentation task. We evaluate the TotalSegmentator (TS) model against our rib segmentation model. We report mean and standard deviation. The arrow after the metric indicates whether higher or lower values are better. The TotalSegmentator cannot segment the beginning part of the rib, which is reflected by these metrics.}
    \label{tab:evaltest}
\end{table}

\begin{table}[hbpt]
    \centering
    \small
    \begin{tabular}{rcccc}
        \toprule
        & RibSeg & RibSeg HF & TS \cite{wasserthal2023totalsegmentator} & Ours \\
        \midrule
        Binary DSC $\uparrow$                & $0.644 \pm 0.059$ & $0.754 \pm 0.056$ & $0.779 \pm 0.017$ & $\textbf{0.999} \pm 0.003$ \\
        $\text{RQ} \uparrow$                 & $0.974 \pm 0.031$ & $0.974 \pm 0.031$ & $0.961 \pm 0.044$ & $\textbf{0.976} \pm 0.033$\\
        $\text{SQ}_{\text{DSC}} \uparrow$    & $0.647 \pm 0.059$ & $0.747 \pm 0.057$ & $0.761 \pm 0.029$ & $\textbf{0.990} \pm 0.020$   \\
        $\text{PQ}_{\text{DSC}} \uparrow$    & $0.630 \pm 0.059$ & $0.728 \pm 0.060$ & $0.732 \pm 0.054$ & $\textbf{0.967} \pm 0.044$\\
        $\text{SQ}_{\text{ASSD}} \downarrow$ & $1.931 \pm 0.485$ & $1.840 \pm 0.483$ & $1.916 \pm 0.478$ & $\textbf{0.071} \pm 0.143$\\
        \bottomrule
    \end{tabular}
    \caption{Evaluation of the 20 manually corrected test subjects from the RibFrac dataset on the rib segmentation task. We evaluate the TS model, the official RibSeg annotations once without and with holes filled (HF), and our rib segmentation model. We report mean and standard deviation. The arrow after the metric indicates whether higher or lower values are better.}
    \label{tab:evaltestribfrac}
\end{table}

\begin{table}[hbpt]
    \centering
    \small
    \begin{tabular}{lr}
        \toprule
        & Percentage \\
        \midrule
        Ribs rated (5) & 69.3 \\ %597
        Ribs rated (4) & 16.7 \\ %144
        Ribs rated (3) & 12.2 \\ %106
        Ribs rated (2) & 1.2 \\ %10
        Ribs rated (1) & 0.6 \\ %5
        \midrule
        Ribs have correct measure & 98.2 \\ %817
        Ribs incorrectly segmented & 1.8 \\ %15
        \bottomrule
    \end{tabular}
    \caption{Quantitative results of the manual review of the lowest ribs across RibFrac and VerSe data. The majority of segmentations were correct or showed only minor errors which did not influence the length measurement algorithm. The percentages are all compared to the total number of ribs analyzed and reviewed. Each block separated by a horizontal line sums to 100 \%. }
    \label{tab:testdata}
\end{table}

\begin{table}[hbpt]
    \centering
    \small
    \begin{tabular}{lcc}
        \toprule
        & \multicolumn{2}{c}{SVM Kernel} \\
        \cmidrule(lr){2-3}
        Input Features & Polynomial & Linear \\
        \midrule
        $2$-PPR & $ 0.53 \pm 0.04 $ & $ 0.38 \pm 0.09 $ \\
        $3$-PPR & $ 0.70 \pm 0.04 $ & $ 0.73 \pm 0.04 $ \\
        $4$-PPR & $ 0.82 \pm 0.04 $ & $ 0.82 \pm 0.04 $ \\
        DRC & $ 0.62 \pm 0.04 $ & $ 0.60 \pm 0.06 $ \\
        \midrule
        $2$-PPR and DRC & $ 0.64 \pm 0.05 $ & $ 0.71 \pm 0.04 $\\
        $3$-PPR and DRC & $ 0.75 \pm 0.05 $ & $ 0.78 \pm 0.05 $\\
        $4$-PPR and DRC & $ 0.79 \pm 0.02 $ & $\textbf{0.84} \pm 0.04$\\
        \bottomrule
    \end{tabular}
    \caption{Evaluation of the SVMs trained on morphological features to classify stump ribs against regular ribs. We report the mean F1-score and standard deviation of SVMs trained with polynomial or linear kernel on different input features over 10 different seeded runs. The value in bold marks the best-performing combination.}
    \label{tab:svmresults}
\end{table}
%%%%%%%%%%%%%%%%%%%%%%%%%%%
\section{Glossary}

\begin{table}[hbpt]
    \centering
    \small
    \begin{tabular}{rl}
        \toprule
        Abbreviation & Description \\
        \midrule
        CT & Computed Tomography \\
        SR & Stump rib \\
        TLTV & Thoracolumbar Transitional Vertebra \\
        RLMA & Rib Length Measurement Algorithm \\
        DRC & Direction Vector between Rib and Vertebra Corpus\\
        PDRC & Posterior Component of the DRC vector \\
        PPR & Path Point Spatial Relation \\
        $n$-PPR & The PPR of the first $n$ path points \\
        TS & TotalSegmentator \\
        DSC & Dice similarity coefficient \\
        ASSD & Average symmetric surface distance \\
        RQ & Recognition Quality \\
        SQ & Segmentation Quality \\
        PQ & Panoptic Quality \\
        IoU & Intersection over Union \\
        SVM & Support Vector Machine \\
        \bottomrule
    \end{tabular}
    \label{tab:glossary}
\end{table}

\section{Funding sources}

This study has received funding from the European Research Council (ERC) under the European Union’s Horizon 2020 research and innovation program (101045128—iBack-epic—ERC2021-COG).

\appendix
\section{Appendix}
\label{app}

\subsection{\CB{Qualitative Failure Analysis}}\label{app:failure}

\REV{Based on our analysis set of 2464 individual ribs across 648 subjects, we had noticeable segmentation errors in 1.8 \% of them.}

\begin{figure}[!htbp]
    \includegraphics[width=\textwidth]{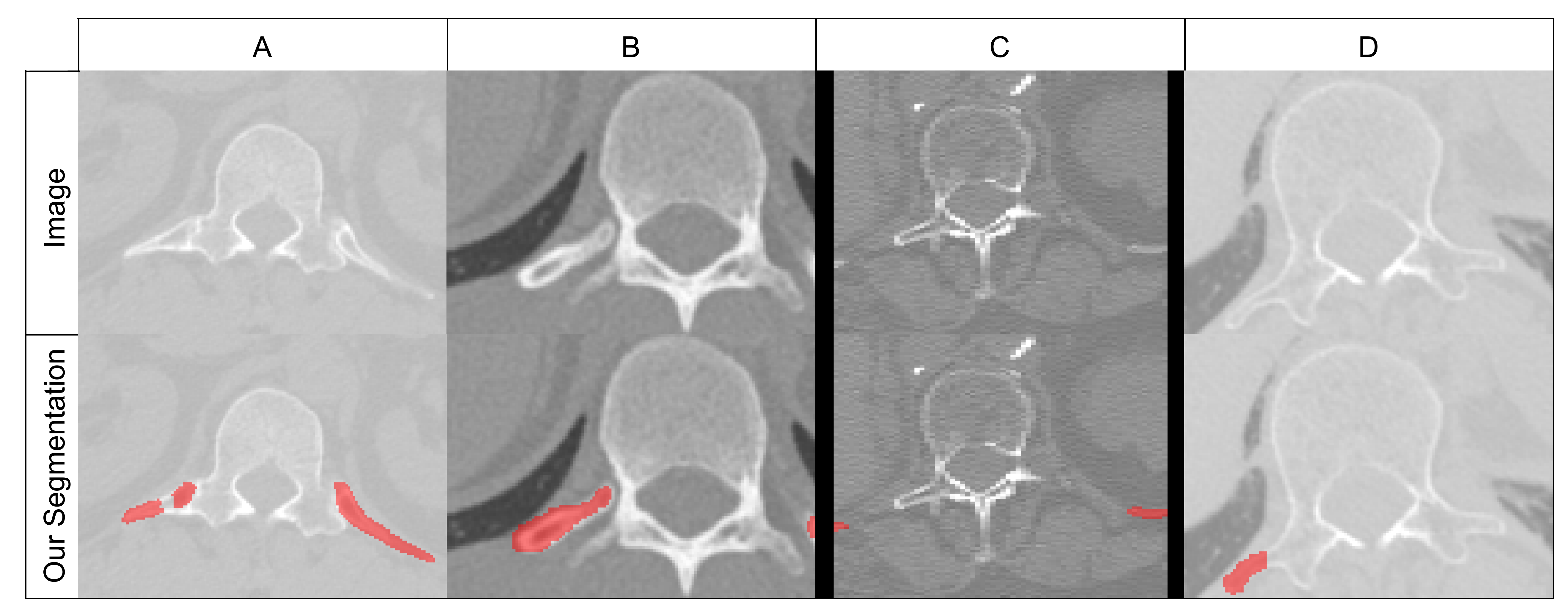}
    \caption{\REV{Examples of challenging and failure cases. The top row portrays the image, and the bottom overlays our predicted segmentation. A) A challenging case where the distinction between costal process and rib is unclear due to fusion. The two components are joined in other slices, and the measured rib length would not have influenced the SR label. B) Imperfect segmentation quality that did not affect the measured rib length. C) A scan that is not wide enough to fully capture the ribs (the left and right black areas are the image's ends). The measured rib length of the visible parts is correct; thus, we know if it is not an SR if that length already exceeds the threshold. D) An error where the segmentation model confused the costal process with a rib and segmented it, detecting a rib where there isn't.}}
    \label{fig:segfail}
\end{figure}

\REV{Figure }\ref{fig:segfail} \REV{demonstrates the typical cases we observed. Most segmentation errors are small and local and do not noticeably influence the measured rib length. The failure cases consist primarily of samples with a fusion; thus, the distinction between the costal process and the rib is rather challenging. For one case, the model segmented the costal process of a lumbar vertebra. We can only hypothesize this is an edge case where the costal process and surrounding image structures resemble a rib.}

\REV{In the future, extracting exact vertebra labels from vertebra labeling approaches and utilizing those to eliminate these rarely segmented costal processes could be an automated way of detecting these errors. Another way could be to calculate normative values across lots of manually checked data and then detect outliers by comparing the features against the normative distribution.}

\subsection{Moving Stump Rib Threshold}\label{app:srthresh}

\begin{figure}[!htbp]
    \centering
    \includegraphics[width=0.80\textwidth]{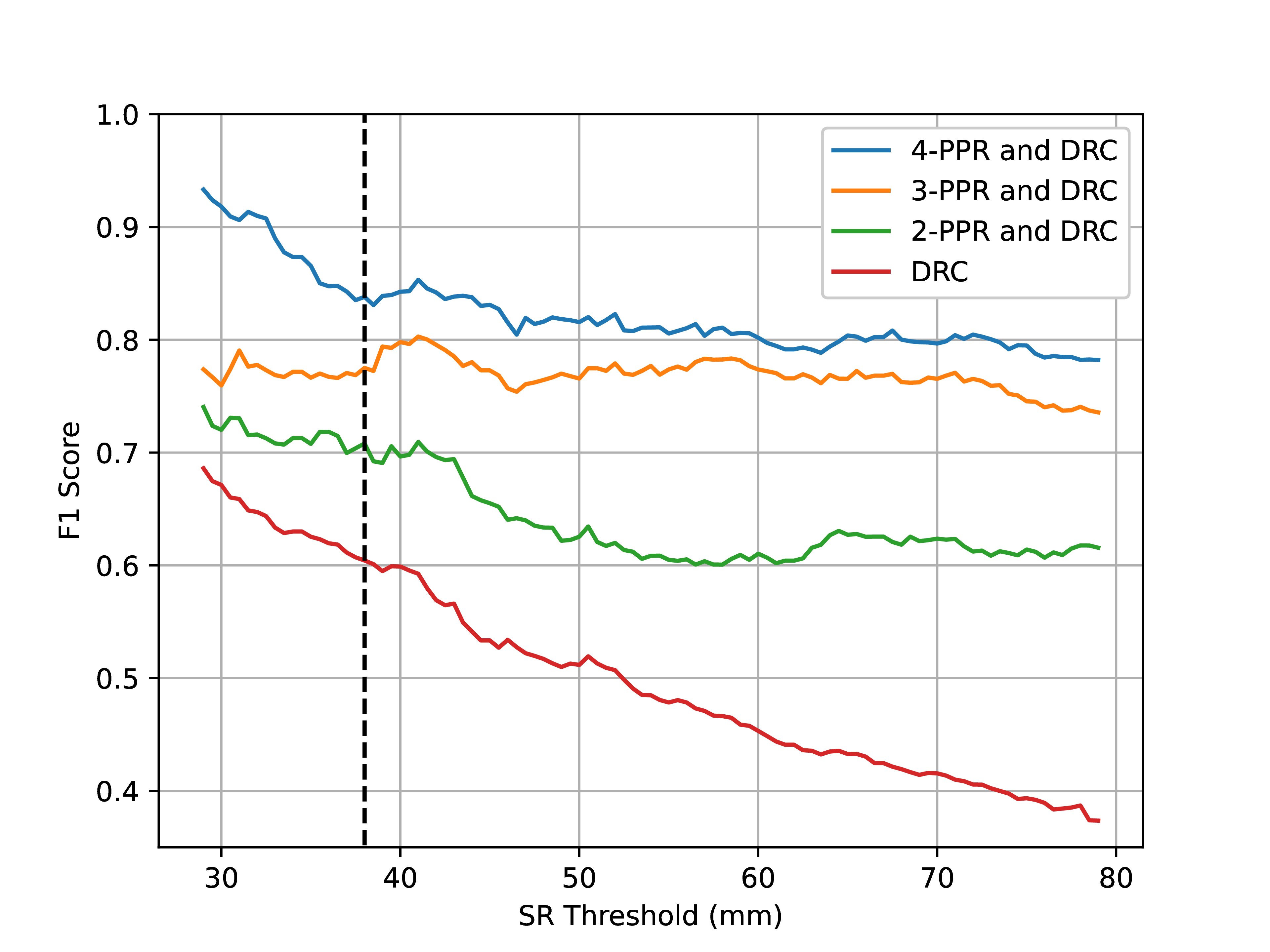}
    \caption{The relation of the classification performance of a linear SVM and the stump rib threshold used to create the reference SR labels. The x-axis shows the SR threshold in mm, while the y-axis is the average F1 score over 10 runs. The dashed black line highlights the generally defined length threshold of $38$ mm.}
    \label{fig:svm_srthres}
\end{figure}

The length threshold defined for stump ribs is $38$ mm. When experimenting with different shifts to this threshold, the F1-score using $4$-PPR and DRC is consistently above all other sets of SVM inputs (see Figure \ref{fig:svm_srthres}). Using three points is still consistently better than using only the first two path points. With an increasingly large threshold, all performance curves seem to decrease but have a small peak at around a chosen threshold of $41$ mm. Please note that an F1-score of $0.5$ means that the two classes (SR and non-SR) are not distinguishable by the features. %This seems to be the case when using only the first two path points, regardless of the chosen SR threshold.

If the stump rib threshold would truly differentiate ribs into categories beyond the length, we would've expected a peak at the length threshold of $38$ mm. This is not the case. In fact, there are mostly no hard edges or peaks but rather fluent changes to the performance. This indicates that either a stump rib labeling purely based on a length threshold is sub-optimal or that our morphological features smoothly relate to the length of the ribs without clear boundaries. There seem to be small peaks at around the threshold of 41 mm, for which we can offer no explanation.

%%%%%%%%%%%%%%%%%%%%%%%%%%%
%% For citations use: 
%%       \cite{<label>} ==> [1]

%%
%Example citation, See \cite{segmentation_survey}.

%% If you have bib database file and want bibtex to generate the
%% bibitems, please use
%%
\bibliographystyle{elsarticle-num} 
\bibliography{bib}

%% else use the following coding to input the bibitems directly in the
%% TeX file.

%% Refer following link for more details about bibliography and citations.
%% https://en.wikibooks.org/wiki/LaTeX/Bibliography_Management
\end{document}